%% file: main.tex
\documentclass[letterpaper, 10 pt, conference]{ieeeconf}  

\IEEEoverridecommandlockouts                              
\usepackage{amssymb}
\usepackage{graphicx,amsmath}
\usepackage{lineno}
\usepackage{caption}
\usepackage{subcaption}
\usepackage{gensymb}

\title{\LARGE \bf
Learning Active Spine Behaviors for Dynamic and Efficient Locomotion in Quadruped Robots
}

\author{Shounak Bhattacharya, Abhik Singla, Abhimanyu, Dhaivat Dholakiya, \\
Shalabh Bhatnagar, Bharadwaj Amrutur, Ashitava Ghosal and Shishir Kolathaya
\thanks{*This work is supported by the Robert Bosch Center for Cyber Physical Systems, Bangalore, India and DST INSPIRE
Faculty Fellowship No. IFA17-ENG212}
\thanks{Shounak Bhattacharya, Abhik Singla, and Dhaivat Dholakiya
are with the Robert Bosch Centre for Cyber-Physical Systems,
IISc, Bangalore, India. E-mail: {\tt\small {\{shounakb, abhiksingla, dhaivatd\}@iisc.ac.in }}}%
\thanks{Abhimanyu is with Department of Mechanical Engineering, BITS Pilani, Goa campus, India E-mail:
{\tt\small {abhimanyusingh8713@gmail.com }}}%
\thanks{Shalabh Bhatnagar is with the Faculty of Computer Science and
Automation, Bharadwaj Amrutur is with the Faculty of Electrical \& Computer Engineering, Ashitava Ghosal is with the Faculty of Mechanical Engineering, and Shishir Kolathaya is an INSPIRE Faculty Fellow at the Robert Bosch Center for Cyber Physical Systems, IISc, Bengaluru, India {\tt\small \{shalabh,amrutur,asitava,shishirk\}@iisc.ac.in}}%
}
\begin{document}
\maketitle
\thispagestyle{empty}
\pagestyle{empty}
\begin{abstract}

In this work, we provide a simulation framework to perform systematic  studies  on  the  effects  of  spinal  joint  compliance and actuation on bounding performance of a $16$-DOF quadruped spined robot Stoch 2.
Fast quadrupedal locomotion with active spine is an extremely hard problem, and involves a complex coordination between the various degrees of freedom. Therefore, past attempts at addressing this problem have not seen much success.
Deep-Reinforcement Learning seems to be a promising approach, after its recent success in a variety of robot platforms, and the goal of this paper is to use this approach to realize the aforementioned behaviors.
With this learning framework, the robot reached a bounding speed of $2.1$ m/s with a maximum Froude number of $2$. 
Simulation results also show that use of active spine, indeed, increased the stride length, improved the cost of transport, and also reduced the natural frequency to more realistic values. 



\end{abstract}

\textbf{Key words:} \textit{Deep-RL, Quadruped, Active spine}

\input{Introduction.tex}

\input{Simulation_design.tex}

\input{D-RL.tex}

\input{Results.tex}
\input{Conclusion.tex}








\bibliographystyle{IEEEtran}
\bibliography{main}

\end{document}

%% file: Introduction.tex
\section{Introduction}\label{Sec:Intro}

Speed, for legged animals, is an important aspect in nature for survival. 
Animals like cheetah, hounds 
etc, use their spine to achieve remarkably high speed gaits. 
Studies on these animals show that the spine movement increases the effective stride length, provides auxiliary power to the legs and helps to harness energy by storing and releasing it. 
From the evolutionary point of view, it is natural to consider the spine as a propulsive engine of the vertebrate body. All of these observations point to the fact that spine is crucial for locomotion.

\begin{figure}[!t]
    \centering
    \includegraphics[width=\linewidth]{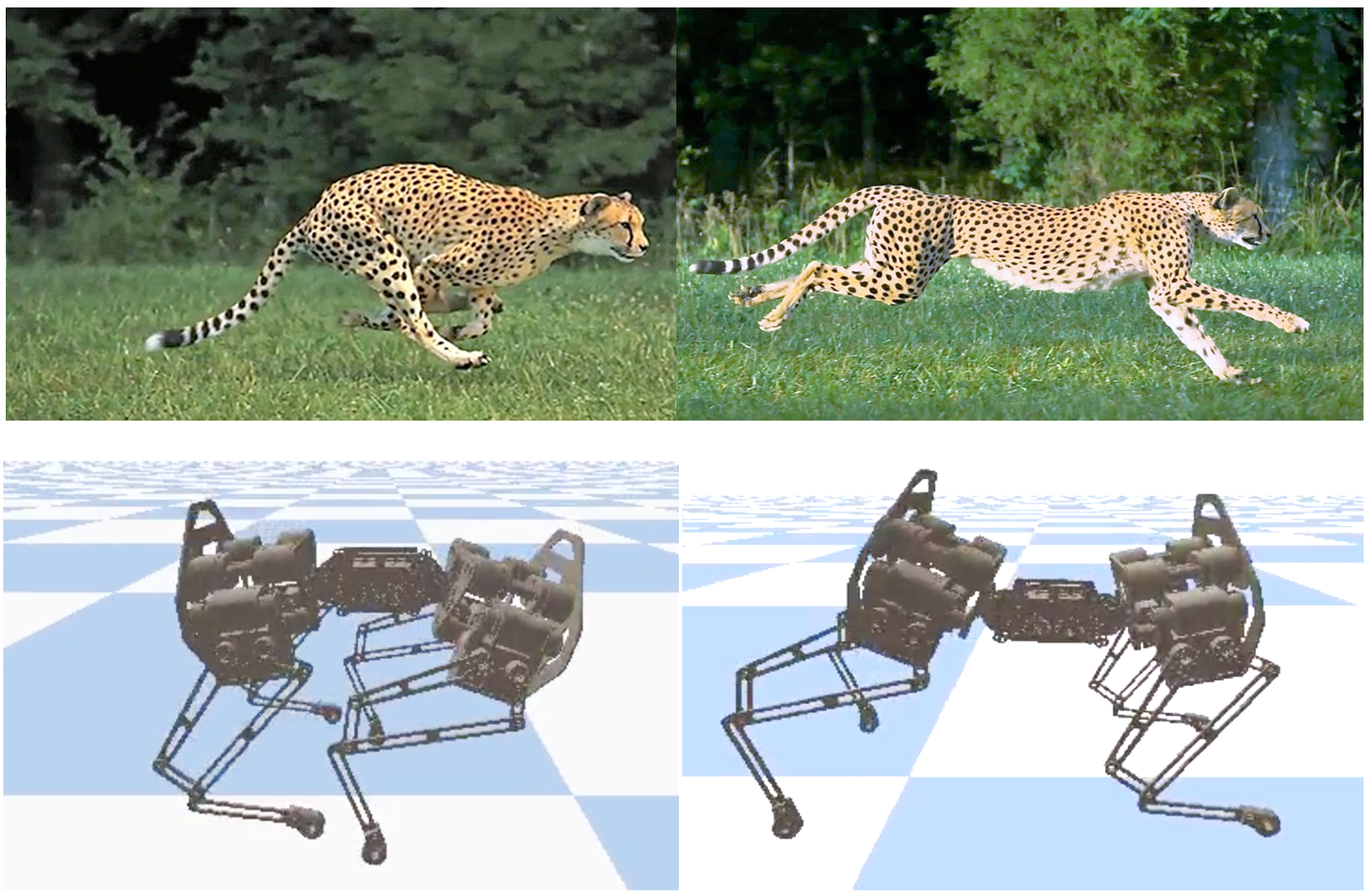}
    \caption{Our simulated robot, Stoch 2, demonstrating active spine motion learned through Deep reinforcement learning in comparison with the Cheetah \cite{Cheetah} with its natural bounding. 
    }
    \label{fig: ComSpine}
\end{figure}

There  have  been quite a few   quadruped  robots  having passive or active spine in their structure, namely, BobCat, Sevel, MIT Cheetah, INU \cite{duperret2017empirical,eckert2018towards, eckert2015comparing, koutsoukis2016passive, chen2017effect}. 
There is a large body of work over planer spine models with one DOF revolute \cite{pouya2017spinal,yesilevskiy2018spine,culha2011quadrupedal},  and prismatic \cite{fisher2017effect} joints.
Apart from the one DOF model, the spines were modeled as point masses in these works.
On the other hand, robots such as Sevel \cite{eckert2018towards}, INU \cite{duperret2017empirical} contain a two DOF spine.
This feature allows the spine to be longer while maintaining its orientation during bounding.
However, reduced order models were used when constructing the empirical model of the robot.
Moreover, in \cite{ duperret2017empirical}, the legs were assumed to be massless, which result in large reality gaps.
From a practical standpoint, we cannot use the approaches mentioned above, due to the fact that the spine models cannot be reduced or approximated, and the legs have significant masses (by upto $25\%$) in the real hardware.
Apart from these, inaccuracy in modeling, imperfect ground contacts play a significant role while transferring on the real hardware. 





To address the above issues and to develop an accurate model of the robot and environment, researchers began to use high fidelity physics engine based simulators \cite{ coumans2013bullet, Webots, TodorovET12} to simulate the robot and environment model.
Works such as \cite{eckert2015comparing, google_paper, chen2017effect, sprowitz2013towards, sprowitz2018oncilla} have pursued this approach to bridge the gap between theory and experiment.
Despite these tools, the results obtained with these simulators still required a noticeable amount of manual tuning and intervention to deploy on hardware.
This is not practically feasible.

\begin{figure*}[t]
    \centering
    \includegraphics[width=0.9\linewidth]{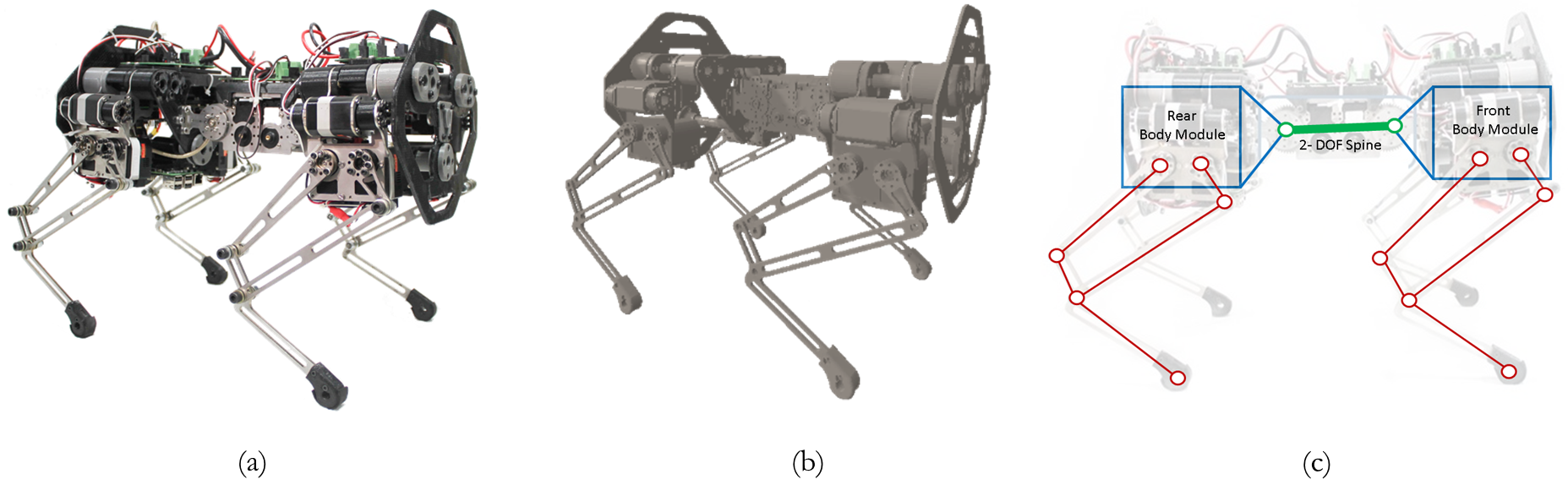}
    \caption{Figure containing the (a) the quadruped robot, Stoch 2, (b) simulated robot model in PyBullet physics engine, and (c) skeleton  of Stoch 2 depicting the front, rear and the spine module.}
    \label{fig:stoch_hw}
\end{figure*}


The principal challenge to be solved in using spine is that of control, namely, determining what actions should be applied over time in order to produce efficient and natural looking gaits. 
It is still an open question as to what type of control strategies can enhance locomotion performance with an active spine. Spine introduces redundancy, and developing a control strategy is often challenging and time consuming.
Complexities arise due to nonlinear and coupled dynamics of legged systems, and from the existing trade-off between different performance criteria such as gait speed, energy efficiency, and stability.
Therefore, most of the existing quadruped robots of today have completely avoided using spine by featuring a single rigid body with four legs with individually actuated hips and/or knees.

We have previously explored Deep Reinforcement learning (D-RL) based methods to realize efficient quadrupedal walking in Stoch \cite{singla2018realizing}, and our goal in this paper is to investigate the efficacy of D-RL for active spine behaviors. In other words, we want to revisit the problem of spine bounding with this new approach, which has seen a lot of success in recent years \cite{google_paper,singla2018realizing,sac,cassie}.
Therefore, in this work, we provide insights into how RL can be successfully applied to such problems.
To the best of our knowledge, this is one of the earliest work which aims to learn complex control policy for a spined quadruped robot.
We also evaluate numerous performance metrics to validate the advantage of spine. Also, we show that the final control policy produces actions that are both robust and efficient.

We have arranged the paper in the following manner.
Section \ref{sec: Sim_des} describes the design of our simulated robot model and simulation framework used for the training.
Section \ref{sec: rl} describes the D-RL framework used to learn the control policy. 
Finally, in section \ref{sec: Results}, all the results obtained through various simulations are discussed thoroughly.

%% file: Simulation_design.tex
\section{Simulation model design} \label{sec: Sim_des}

\textit{Stoch 2}  is  a  quadrupedal  robot designed  and developed in-house at the Indian Institute of Science (IISc), Bangalore, India.
It is the second generation robot in the \textit{Stoch} series. 
A detailed description of the previous model can be found in  \cite{singla2018realizing, dhaivatdesigndevelopment}.
In this section, we first describe the overview of Stoch 2 robot and the essential hardware details, and then describe the simulation framework used for training. The scope of this work is limited to the simulation results portraying quadrupedal spine locomotion, whereas the hardware experiments are a part of our future work.

The robot is designed as three modules: two body modules and one spine module. The body modules are connected via spine, as shown in Fig. \ref{fig:stoch_hw}a.
The overall size and form factor of the robot is similar to \textit{Stoch} \cite{dhaivatdesigndevelopment}.
Each body module is composed of two legs. 
A single leg contains three degrees of freedom. 
Each of them corresponds to the flexor and extension movement of hip, knee and abduction movements. However, the simulation model uses only hip and knee motion while keeping the abduction locked in position.
Each leg comprises of a five bar linkage mechanism, where two of the links remain actuated. This enables the leg to follow a given trajectory in a plane. 
The central spine is designed as a serial $2$ DOF mechanism. Each of the spine joint is actuated.
Overall, the robot simulation model consists of $10$ actuated degrees of freedom, including four legs and the spine. 
The key specifications of the simulation model are summarized in Table \ref{tab:robot_specs}.

\begin{table}[h!]
 \centering
 \begin{tabular}{l r}
 \hline
 \textbf{Parameter} & \textbf{Value}\\
 \hline
spine length & $102$ mm\\ 
total leg length & $245$ mm\\
min./max. hip joint angle & $-45^{\circ}$/ $45^{\circ}$\\
min./max. knee joint angle & $-70^{\circ}$/ $70^{\circ}$\\
min./max. spine front joint angle & $-15^{\circ}$/ $15^{\circ}$\\
min./max. spine back joint angle& $-15^{\circ}$/ $15^{\circ}$\\

 \hline
 \end{tabular}
 \caption{Table showing the details of the simulated robot in PyBullet.}
 \label{tab:robot_specs}
 \end{table}

\begin{figure}[t!]
    \centering
    \includegraphics[width = \linewidth]{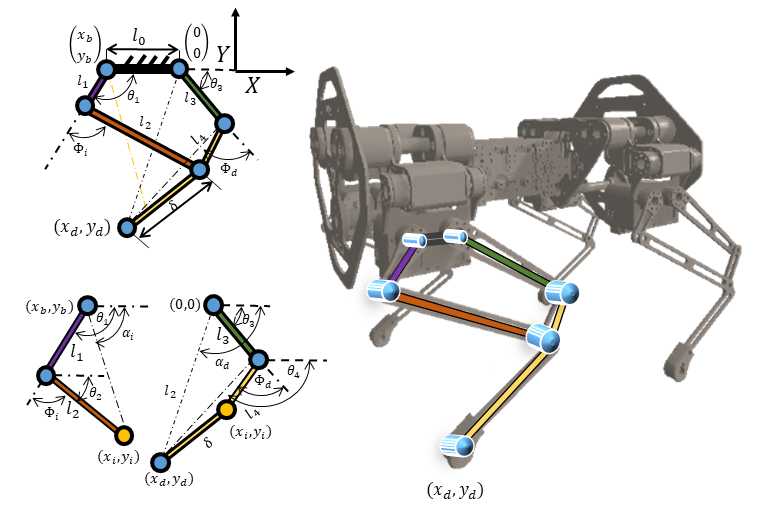}
    \caption{Detailed kinematic model for the front left leg is shown here.}
    \label{fig:IK_S2}
\end{figure}

\begin{figure*}[t!]
\centering
\includegraphics[width = \linewidth] {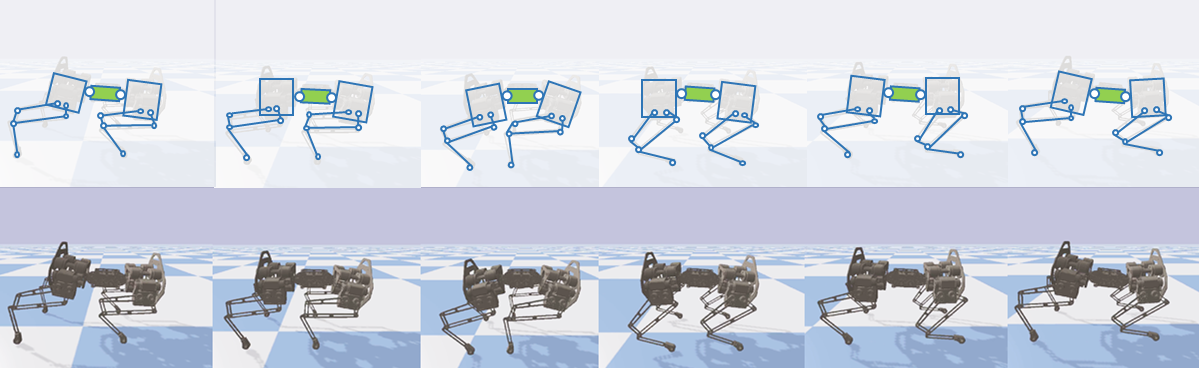}
\caption{Figure showing the tiles of the simulated model showing the learned bound gait with active spine. The top tiles show the skeleton and the bottom tiles show the robot in PyBullet simulator. }
\label{fig:tiles}
\end{figure*}

We used Pybullet \cite{pybullet} simulator, built on top of Bullet3 physics engine. A three-dimensional computer-aided-design (CAD) model is developed using SolidWorks \cite{solidworks} to capture the kinematics and inertial properties of the robot. This model is transferred to Pybullet by using a Universal Robot Description Format \cite{roswebsite}
 (URDF) exporter. In addition, actual mass of all the links, actuator force limits, joint limits and kinematic-loop constraints of the flexural joints are measured and manually updated in the URDF file for a more realistic simulation.


%% file: D-RL.tex
\section{Reinforcement Learning based Controller}\label{sec: rl}

In this section, we will outline the deep reinforcement learning framework used for learning spine based locomotion behaviours.

\subsection{Background}
In reinforcement learning setting, the locomotion problem is formulated as a Markov Decision Process (MDP). An MDP is characterized by a tuple $\{S, A, P, R, \gamma \}$ where $S \subset \mathbb{R}^n$ is the set of robot states referred to as state space, and $A \subset \mathbb{R}^m$ is the set of feasible actions referred to as the action space. The transition probability function $P: S\times A \times S\rightarrow [0,1]$  models the evolution of states based on actions, and $R: S \times A \rightarrow \mathbb{R}$ is the scalar value at every transition step. $\gamma$ is called the discount factor defined in the range $(0,1)$. 

In reinforcement learning, the fundamental idea is to discover a policy, denoted as $\pi: S \times A \to [0,1]$, that maximizes the expected cumulative reward over time. A parameterized policy $\pi_\theta$ with the parameters $\theta$ is the probability density of $a_t$ given $s_t$. The optimal parameters of the policy yield the maximum sum of the cumulative rewards given by
\begin{equation}
    \max_\theta J(\theta) = E\Bigg [\sum\limits_{t=0}^{T} \gamma^t r(s_t, a_t) \Bigg]
\end{equation}
Policy gradients \cite{peters2006policy} is one of the popular methods to solve this optimization problem which takes gradient ascent steps in the direction of increasing cumulative reward. We discuss more details about the algorithm used in Section \ref{algo}.


\subsection{State and Action Space}

\label{sec:model}

\subsubsection{State Space}
Similar to the work in \cite{singla2018realizing}, the state is represented by angles, velocities, torques of the active joints (legs and spine), and body orientation (in quaternions). The combined representation yields a 34-D state space. 
\subsubsection{Action Space}
 The action space consists of the continuous-valued control signal for each active joint. For each leg, the agent learns the legs' end-point positions in polar coordinates represented as $\{r_i, \alpha_i\}$ where $i \in \{1,2\}$. This particular choice of action space ensures that the five-bar leg mechanism never encounters a singularity. We use a custom inverse kinematics solver to compute the joint angles from polar coordinates. 
 As seen from Fig \ref{fig:IK_S2}, the five bar linkage is divided into two 2R Serial linkage and solved for each branch. The details of the equations for a 2R Serial linkage can be found here \cite{dhaivatdesigndevelopment}. Safety limits are also included in the inverse kinematic solver to avoid singular positions.
 However, the agent directly learns the joint angle $\{\beta\}$ for the spine motors. The two motors in the spine are coupled with the relation $\beta_{rear} = -\beta_{front}$ where $\beta$ represents the joint angle. During the bound gait, we learn separate end-point trajectories for front and rear leg pairs. Note that both the legs in front and back module executes the same end-point trajectory during bound. The polar coordinates for the four legs and joint angle for the spine collectively provide a $5$ dimensional action space. 
The polar coordinates and the spine angle are restricted to a bounding box, thereby indirectly imposing angle limits on the joint angles.

\subsection{Reward Function}
We designed a reward function that gives more positive reinforcement as the robot's base speed attains a desired velocity, and simultaneously penalizes high energy consumption. The agent receives a scalar reward after each action step according to the reward function
\begin{align}\label{eq:reward}
r_t = w_{vel} \cdot \mathcal{N}(v_{des},\,\sigma^{2}) \cdot {\sigma \sqrt {2\pi } } - w_E\cdot \Delta E .
\end{align}

Here $v_{des}$ is the desired velocity along the $x$-axis, and $\sigma$ is manually adjusted for various values of $v_{des}$. 
$\Delta E$ is the energy spent by the actuators for the current step,
and $w_{vel}$, $w_E$ are the weights corresponding to each term ($1.0$ and $0.02$ respectively in our experiments). $\Delta E$ is computed as
\begin{align}
\Delta E = \Sigma_{i=1}^{10} ( | \tau_i (t) \cdot \omega_i (t) |  ) \cdot \Delta t,
\end{align}
where $\tau_i$ are the motor torques, and $\omega_i$ are the motor velocities of the $i^{th}$ motor respectively.

\subsection{Network and Learning Algorithm} \label{algo}
We are employing Proximal Policy Optimization (PPO) \cite{PPO} for learning optimal polices that yield action values in continuous space \cite{singla2018realizing}, \cite{google_paper}. 
The actor and critic network in the learning algorithm consists of two fully connected layers with the first and second layers consisting of 128 and 64 nodes respectively. Activation units in the actor and critic networks are ReLU $\rightarrow$ ReLU $\rightarrow$ $\tanh$, and ReLU $\rightarrow$ ReLU $\rightarrow$ linear respectively. 
We used the open source implementation of PPO by Tensorflow Agents \cite{tf_agents} that creates the symbolic representation of the computation graph. The implementation is highly parallelized and performs full-batch gradient ascent updates, using Adam \cite{adam} optimizer, on the batch of data collected from multiple environment instances. The pybullet simulator is configured to spawn multiple agents for faster collection of samples and in our case 30 agents were used in parallel threads.
The proposed approach yields walking gaits that are efficient, fast and also robust in simulation. The learning algorithm runs for a maximum of $10$ million steps and the observed training time is $4$ hours on a Intel Core i7 @3.5Ghz$\times 12$ cores and 32 GB RAM machine.

Having obtained an optimal policy for bounding, we will now discuss the simulation results and also justify how the use of active spine is beneficial to locomotion.

%% file: Results.tex
\section{Results} \label{sec: Results}

\begin{figure}[h!]
    \centering
    \includegraphics[width =\linewidth] {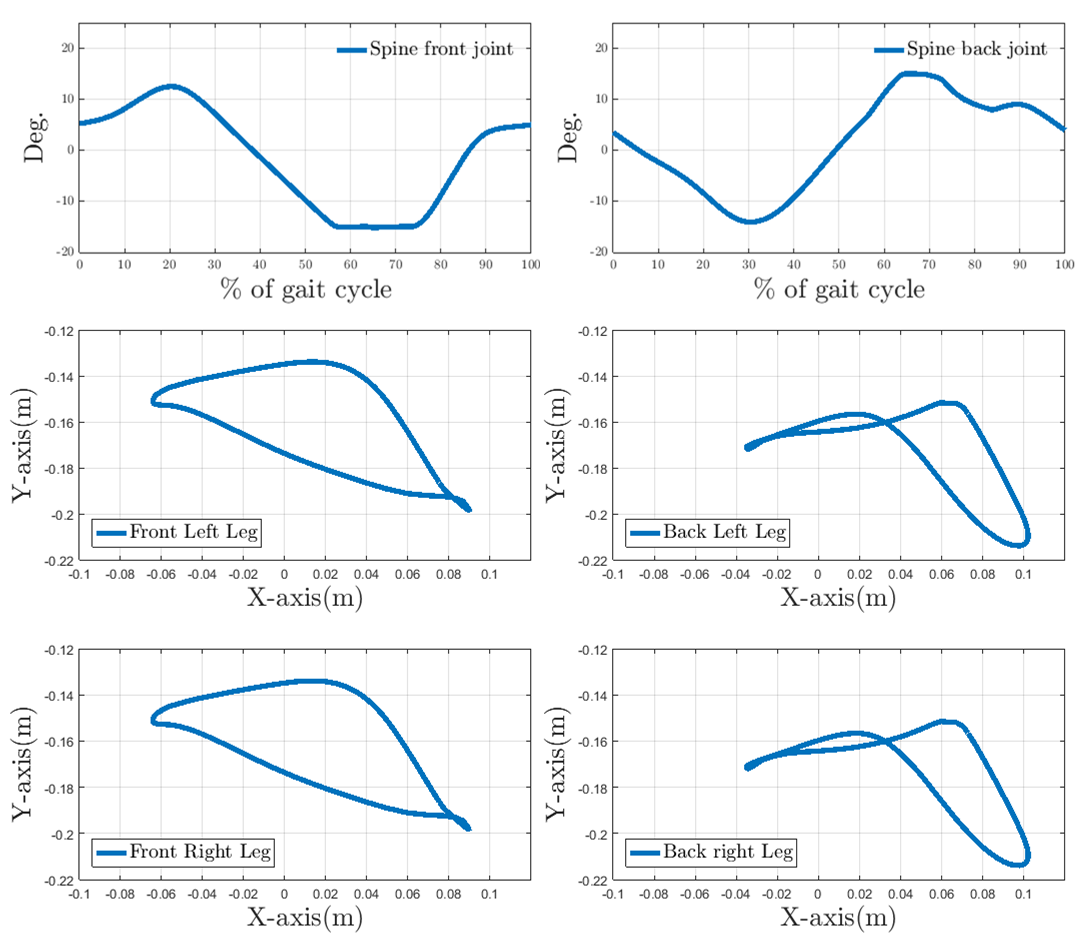}
    \caption{Joint angle of spine and robot's end point trajectory are shown here.}
    \label{fig:endpointtrajectory}
\end{figure}

In this section, we will provide simulation results and also make comparisons between spine and rigid models. We trained multiple bounding gaits with target speeds ranging from $0.5$ m/s and $2.5$m/s for both the models. The maximum speed obtained was $2.21$m/s with an active spine, and $1.8$m/s without spine. Figure \ref{fig:tiles} shows the gait tiles for the target speed $2$m/s. 
Froude number\footnote{Froude number $F_r$ is the ratio of square of the maximum velocity over the gravity and leg length, i.e., $F_r = {v^2}/{g l_0}$, $v$ is the velocity, $l_0$ is the leg length.} 
observed was $2$ at $5.5$ body lengths per second. Figure \ref{fig:endpointtrajectory} shows the end point trajectories and spine angles for the same gait. We will analyze the gaits obtained based on traits like cost of transport, stride lengths, power, torque profiles etc.
Video results are provided in the following link: \url{https://youtu.be/INp4aa-8z2E}.


\begin{figure}[h!]
\centering
\includegraphics[width =0.8\linewidth] {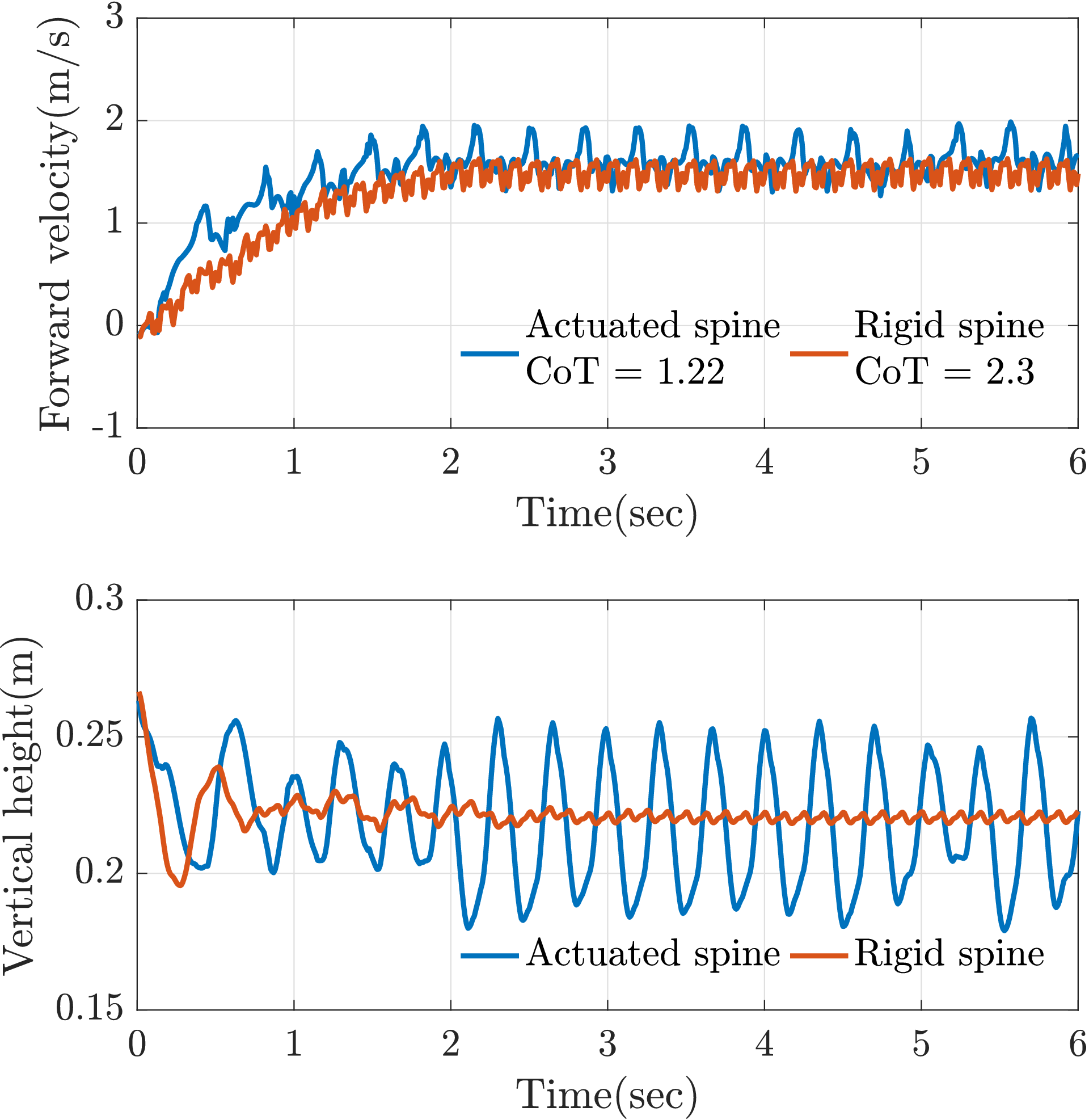}
\caption{Forward velocity and the vertical displacement of the center of mass data obtained for bound gait are shown here. The CoT for the rigid model is almost two times higher for the same speed.}
\label{fig:For_Ver}
\end{figure}
\subsection{Cost of Transport (CoT)}

\begin{figure}[h!]
    \centering
    \includegraphics[width =\linewidth] {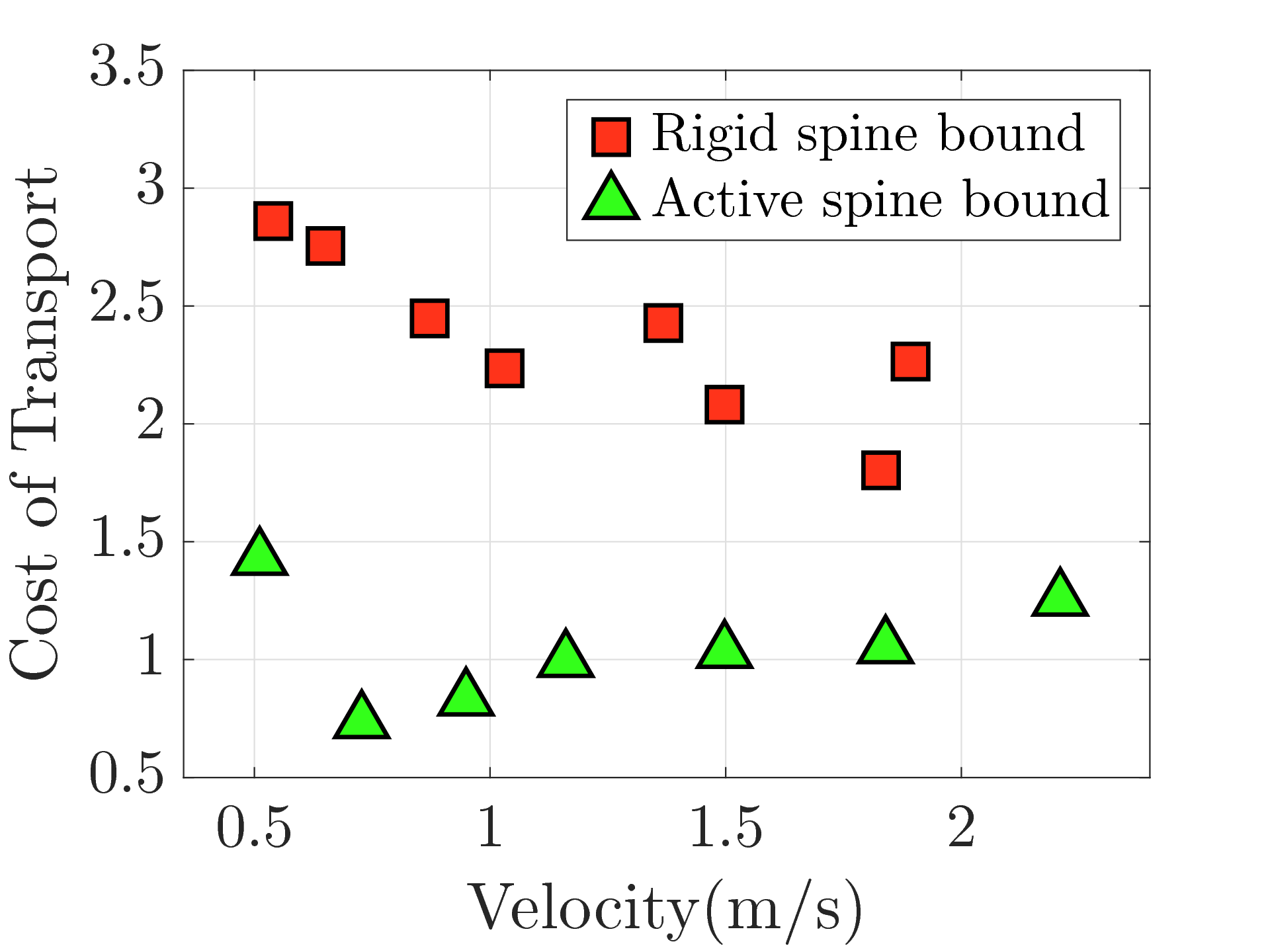}
    \caption{Cost of transport (CoT) vs. forward velocity for the bound gait is shown here. The CoT for the rigid model is almost two times higher than that for the spine model.}
    \label{fig:CoT_Velocity}
\end{figure}

\begin{figure}[h!]
    \centering
    \includegraphics[width = 0.75\linewidth]{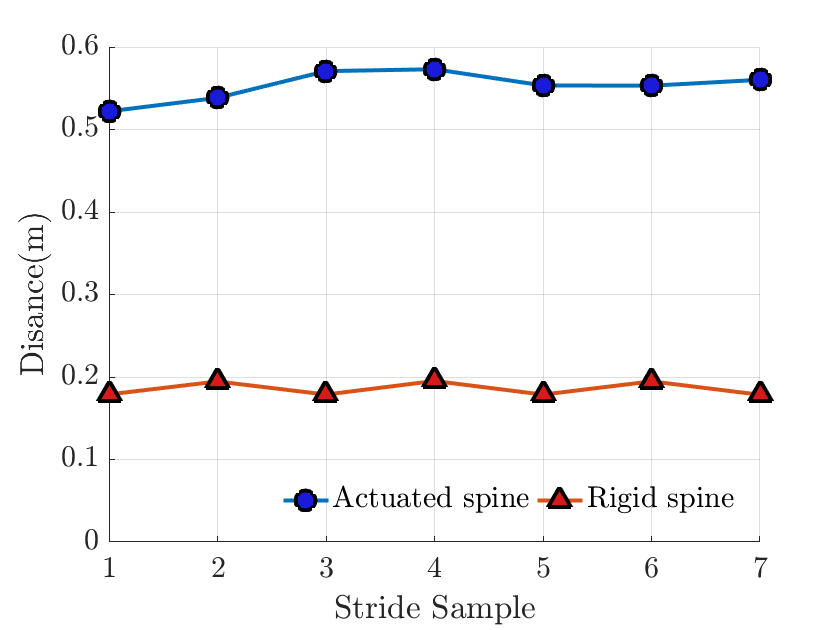}
    \caption{Comparing stride length between an active spine and a rigid spine bound, with a forward velocity of $1.5$m/s. The spine bound shows a much higher stride length.}
    \label{fig:stride_len}
\end{figure}

In our experiments, we use Cost of Transport (CoT) as a metric of measurement for the gait performance.  
CoT is nothing but the mechanical work normalized by the weight and the distance traveled.
To determine CoT, we take the integral of the positive work done by the actuators \cite{pouya2017spinal}:
$$CoT = \dfrac{\int_{t = 0}^{T} \max (\Gamma(t) . \dot{\theta}(t) , 0) dt}{(M.g.\Bar{X})},$$
where $\Gamma(t)$ is the torque of the motors. 
$\dot{\theta}$ is speed of the motor.
$\Bar{X}$ is the average distance covered by the robot along X-axis. 
$M$ is the mass of the robot.
$g$ is the gravity.

We pick $5$ trials and then average the CoT computed over these trials for consistency. 
Figure \ref{fig:For_Ver} shows the velocity profiles and center of mass height of the spine and rigid models. Fig. \ref{fig:CoT_Velocity} shows the CoT comparison between the rigid and spine models. It can be verified that CoT of the spine model is always less than that of the rigid model for the velocities shown. Moreover, the CoT for rigid models is almost twice as much as that for the spine models.
This observation is concurrent with the results obtained in \cite{khoramshahi2013benefits, culha2011quadrupedal, pouya2017spinal, yesilevskiy2018spine}. 

\subsection{Stride Length Comparison}
Stride length provides a measure of how far the robot has walked during each step.
It is the distance between two successive placements of the same foot. To measure the stride length, heel to heel distance is determined. 
It can be verified from Figure \ref{fig:stride_len} that there is a $ \approx200\% $ increase in stride length due to active spine.  
Moreover, spine helps in reducing the bounding frequency \cite{culha2011quadrupedal}, thereby allowing the gaits to be more realistic.

\subsection{Torque, Power Profile and Gait Diagram}

\begin{figure}[h!]
    \centering
    \includegraphics[width =\linewidth]{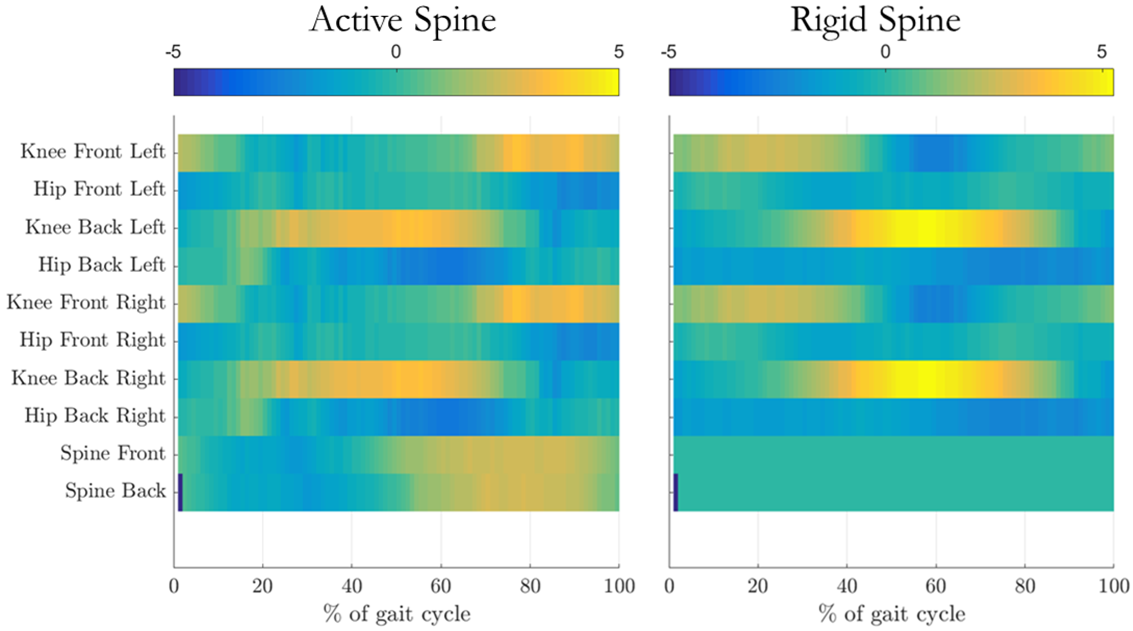}
    \caption{Figure showing the torque profile for active and rigid spine over a complete gait cycle.}
    \label{fig:Torque_graph}
\end{figure}

In figure \ref{fig:Torque_graph}, we compare the torque profile for one gait cycle of the bound gait with both active and rigid models, which are at nearly the same speeds.
It can be observed that the peak torque values have depreciated for the active spine by nearly $ 30 \% $. 
Apart from this, it can be verified that the torque profile correlates well with the stance and swing phase of the robot. 
The average torque experienced by the front leg motors are higher, while the impulses experienced by the back legs were higher. 

%% file: Conclusion.tex
\section{Conclusion} \label{sec: Conclusion}

We showed that spine is a key driving factor for the improvement of speed and COT in quadrupeds. Unlike existing approaches, we used Deep-Reinforcement Learning to realize spine behaviors. It is worth noting that the training was from scratch, and did not require prior understanding of the mechanisms involved in using the spine. We have made comparisons with rigid models, and showed that spine improves cost of transport, power, torque, and stride lengths significantly.
Future work will involve implementation of these policies on hardware. 

%% file: main.bbl
\begin{thebibliography}{10}
\providecommand{\url}[1]{#1}
\csname url@samestyle\endcsname
\providecommand{\newblock}{\relax}
\providecommand{\bibinfo}[2]{#2}
\providecommand{\BIBentrySTDinterwordspacing}{\spaceskip=0pt\relax}
\providecommand{\BIBentryALTinterwordstretchfactor}{4}
\providecommand{\BIBentryALTinterwordspacing}{\spaceskip=\fontdimen2\font plus
\BIBentryALTinterwordstretchfactor\fontdimen3\font minus
  \fontdimen4\font\relax}
\providecommand{\BIBforeignlanguage}[2]{{%
\expandafter\ifx\csname l@#1\endcsname\relax
\typeout{** WARNING: IEEEtran.bst: No hyphenation pattern has been}%
\typeout{** loaded for the language `#1'. Using the pattern for}%
\typeout{** the default language instead.}%
\else
\language=\csname l@#1\endcsname
\fi
#2}}
\providecommand{\BIBdecl}{\relax}
\BIBdecl

\bibitem{Cheetah}
G.~Wilson, ``Cheetahs on the edge—director’s cut,''
  \url{https://vimeo.com/53914149}, March 2013.

\bibitem{duperret2017empirical}
J.~Duperret and D.~Koditschek, ``Empirical validation of a spined
  sagittal-plane quadrupedal model,'' 2017.

\bibitem{eckert2018towards}
P.~Eckert, A.~E. Schmerbauch, T.~Horvat, K.~S{\"o}hnel, M.~S. Fischer,
  H.~Witte, and A.~J. Ijspeert, ``Towards rich motion skills with the
  lightweight quadruped robot serval-a design, control and experimental
  study,'' in \emph{International Conference on Simulation of Adaptive
  Behavior}.\hskip 1em plus 0.5em minus 0.4em\relax Springer, 2018, pp. 41--55.

\bibitem{eckert2015comparing}
P.~Eckert, A.~Spr{\"o}witz, H.~Witte, and A.~J. Ijspeert, ``Comparing the
  effect of different spine and leg designs for a small bounding quadruped
  robot,'' in \emph{Robotics and Automation (ICRA), 2015 IEEE International
  Conference on}.\hskip 1em plus 0.5em minus 0.4em\relax IEEE, 2015, pp.
  3128--3133.

\bibitem{koutsoukis2016passive}
K.~Koutsoukis and E.~Papadopoulos, ``On passive quadrupedal bounding with
  translational spinal joint,'' in \emph{Intelligent Robots and Systems (IROS),
  2016 IEEE/RSJ International Conference on}.\hskip 1em plus 0.5em minus
  0.4em\relax IEEE, 2016, pp. 3406--3411.

\bibitem{chen2017effect}
D.~Chen, N.~Li, H.~Wang, and L.~Chen, ``Effect of flexible spine motion on
  energy efficiency in quadruped running,'' \emph{Journal of Bionic
  Engineering}, vol.~14, no.~4, pp. 716--725, 2017.

\bibitem{pouya2017spinal}
S.~Pouya, M.~Khodabakhsh, A.~Spr{\"o}witz, and A.~Ijspeert, ``Spinal joint
  compliance and actuation in a simulated bounding quadruped robot,''
  \emph{Autonomous Robots}, vol.~41, no.~2, pp. 437--452, 2017.

\bibitem{yesilevskiy2018spine}
Y.~Yesilevskiy, W.~Yang, and C.~D. Remy, ``Spine morphology and energetics: how
  principles from nature apply to robotics,'' \emph{Bioinspiration \&
  biomimetics}, vol.~13, no.~3, p. 036002, 2018.

\bibitem{culha2011quadrupedal}
U.~Culha and U.~Saranli, ``Quadrupedal bounding with an actuated spinal
  joint,'' in \emph{Proceedings-IEEE International Conference on Robotics and
  Automation}, 2011, pp. 1392--1397.

\bibitem{fisher2017effect}
C.~Fisher, S.~Shield, and A.~Patel, ``The effect of spine morphology on rapid
  acceleration in quadruped robots,'' in \emph{Intelligent Robots and Systems
  (IROS), 2017 IEEE/RSJ International Conference on}.\hskip 1em plus 0.5em
  minus 0.4em\relax IEEE, 2017, pp. 2121--2127.

\bibitem{coumans2013bullet}
E.~Coumans \emph{et~al.}, ``Bullet physics library,'' \emph{Open source:
  bulletphysics. org}, vol.~15, no.~49, p.~5, 2013.

\bibitem{Webots}
\BIBentryALTinterwordspacing
Webots, ``http://www.cyberbotics.com,'' 2009, commercial Mobile Robot
  Simulation Software. [Online]. Available: \url{http://www.cyberbotics.com}
\BIBentrySTDinterwordspacing

\bibitem{TodorovET12}
\BIBentryALTinterwordspacing
E.~Todorov, T.~Erez, and Y.~Tassa, ``Mujoco: A physics engine for model-based
  control.'' in \emph{IROS}.\hskip 1em plus 0.5em minus 0.4em\relax IEEE, 2012,
  pp. 5026--5033. [Online]. Available:
  \url{http://dblp.uni-trier.de/db/conf/iros/iros2012.html}
\BIBentrySTDinterwordspacing

\bibitem{google_paper}
\BIBentryALTinterwordspacing
J.~Tan, T.~Zhang, E.~Coumans, A.~Iscen, Y.~Bai, D.~Hafner, S.~Bohez, and
  V.~Vanhoucke, ``Sim-to-real: Learning agile locomotion for quadruped
  robots,'' \emph{CoRR}, vol. abs/1804.10332, 2018. [Online]. Available:
  \url{http://arxiv.org/abs/1804.10332}
\BIBentrySTDinterwordspacing

\bibitem{sprowitz2013towards}
A.~Spr{\"o}witz, A.~Tuleu, M.~Vespignani, M.~Ajallooeian, E.~Badri, and A.~J.
  Ijspeert, ``Towards dynamic trot gait locomotion: Design, control, and
  experiments with cheetah-cub, a compliant quadruped robot,'' \emph{The
  International Journal of Robotics Research}, vol.~32, no.~8, pp. 932--950,
  2013.

\bibitem{sprowitz2018oncilla}
A.~T. Sprowitz, A.~Tuleu, M.~Ajaoolleian, M.~Vespignani, R.~Moeckel, P.~Eckert,
  M.~D'Haene, J.~Degrave, A.~Nordmann, B.~Schrauwen \emph{et~al.}, ``Oncilla
  robot: a versatile open-source quadruped research robot with compliant
  pantograph legs,'' \emph{Frontiers in Robotics and AI}, vol.~5, p.~67, 2018.

\bibitem{singla2018realizing}
A.~Singla, S.~Bhattacharya, D.~Dholakiya, S.~Bhatnagar, A.~Ghosal, B.~Amrutur,
  and S.~Kolathaya, ``Realizing learned quadruped locomotion behaviors through
  kinematic motion primitives,'' \emph{arXiv preprint arXiv:1810.03842}, 2018.

\bibitem{sac}
\BIBentryALTinterwordspacing
T.~Haarnoja, A.~Zhou, K.~Hartikainen, G.~Tucker, S.~Ha, J.~Tan, V.~Kumar,
  H.~Zhu, A.~Gupta, P.~Abbeel, and S.~Levine, ``Soft actor-critic algorithms
  and applications,'' \emph{CoRR}, vol. abs/1812.05905, 2018. [Online].
  Available: \url{http://arxiv.org/abs/1812.05905}
\BIBentrySTDinterwordspacing

\bibitem{cassie}
\BIBentryALTinterwordspacing
Z.~Xie, P.~Clary, J.~Dao, P.~Morais, J.~W. Hurst, and M.~van~de Panne,
  ``Iterative reinforcement learning based design of dynamic locomotion skills
  for cassie,'' \emph{CoRR}, vol. abs/1903.09537, 2019. [Online]. Available:
  \url{http://arxiv.org/abs/1903.09537}
\BIBentrySTDinterwordspacing

\bibitem{dhaivatdesigndevelopment}
\BIBentryALTinterwordspacing
D.~Dholakiya, S.~Bhattacharya, A.~Gunalan, A.~Singla, S.~Bhatnagar, B.~Amrutur,
  A.~Ghosal, and S.~Kolathaya, ``Design, development and experimental
  realization of a quadrupedal research platform: Stoch,'' \emph{CoRR}, vol.
  abs/1901.00697, 2019. [Online]. Available:
  \url{http://arxiv.org/abs/1901.00697}
\BIBentrySTDinterwordspacing

\bibitem{pybullet}
``Pybullet:,'' \url{https://pybullet.org/wordpress/}.

\bibitem{solidworks}
``{Solidworks}:,'' \url{https://www.solidworks.com}.

\bibitem{roswebsite}
``{URDF}:,'' \url{https://wiki.ros.org/urdf}.

\bibitem{peters2006policy}
J.~Peters and S.~Schaal, ``Policy gradient methods for robotics,'' in
  \emph{2006 IEEE/RSJ International Conference on Intelligent Robots and
  Systems}.\hskip 1em plus 0.5em minus 0.4em\relax IEEE, 2006, pp. 2219--2225.

\bibitem{PPO}
\BIBentryALTinterwordspacing
J.~Schulman, F.~Wolski, P.~Dhariwal, A.~Radford, and O.~Klimov, ``Proximal
  policy optimization algorithms,'' \emph{CoRR}, vol. abs/1707.06347, 2017.
  [Online]. Available: \url{http://arxiv.org/abs/1707.06347}
\BIBentrySTDinterwordspacing

\bibitem{tf_agents}
D.~Hafner, J.~Davidson, and V.~Vanhoucke, ``Tensorflow agents: Efficient
  batched reinforcement learning in tensorflow,'' \emph{arXiv preprint
  arXiv:1709.02878}, 2017.

\bibitem{adam}
\BIBentryALTinterwordspacing
D.~P. Kingma and J.~Ba, ``Adam: {A} method for stochastic optimization,''
  \emph{CoRR}, vol. abs/1412.6980, 2014. [Online]. Available:
  \url{http://arxiv.org/abs/1412.6980}
\BIBentrySTDinterwordspacing

\bibitem{khoramshahi2013benefits}
M.~Khoramshahi, A.~Sprowitz, A.~Tuleu, M.~N. Ahmadabadi, and A.~Ijspeert,
  ``Benefits of an active spine supported bounding locomotion with a small
  compliant quadruped robot,'' in \emph{Proceedings of 2013 IEEE International
  Conference on Robotics and Automation}, no. EPFL-CONF-186299, 2013.

\end{thebibliography}
